\documentclass{article}

\usepackage{arxiv}

\usepackage[utf8]{inputenc}
\usepackage[T1]{fontenc}
\usepackage{hyperref}
\usepackage{url}
\usepackage{booktabs}
\usepackage{amsmath,amssymb,amsfonts}
\usepackage{nicefrac}
\usepackage{microtype}
\usepackage{graphicx}
\usepackage{algorithmic}
\usepackage{textcomp}

\hypersetup{
  pdftitle={Corpus Augmentation for Sign Language Translation via LLM-Guided Video Stitching},
  pdfsubject={cs.CV, cs.CL},
  pdfauthor={Zsolt Robotka, Ádám Rák, Jalal Al-Afandi, András Horváth, György Cserey},
  pdfkeywords={data augmentation, gloss-free sign language translation, large language models, sign stitching, synthetic training data, vision-language pretraining},
}

\title{Corpus Augmentation for Sign Language Translation via LLM-Guided Video Stitching}

\author{
  Zsolt Robotka\thanks{Corresponding author. E-mail: \texttt{zsolt@deepsign.ai}} \\
  \And
  Ádám Rák \\
  \And
  Jalal Al-Afandi\\
  \And
  András Horváth \\
  \And
  György Cserey \\[0.5em]
  \normalsize Peter Pazmany Catholic University,
  Faculty of Information Technology and Bionics, Budapest, Hungary \\
  \normalsize \textit{(Z. Robotka also with DeepSign Technologies Ltd., Budapest, Hungary)}
}

\renewcommand{\shorttitle}{Corpus Augmentation for SLT via LLM-Guided Video Stitching}
\date{}

\begin{document}
\maketitle

\begin{abstract}
Sign language translation (SLT) converts sign language video into
spoken language text and holds significant promise
for improving accessibility and enabling communication between signing
and non-signing communities.
While large weakly-aligned datasets have enabled pre-training at
scale and gloss-free methods have reduced reliance on expert
annotation, high-quality parallel sign video--text pairs for
fine-tuning remain scarce, limiting generalisation on long-tail
vocabulary and unseen constructions.
We propose a corpus augmentation approach that requires no additional human annotation, external sign-language video corpora, or generative video models, relying only on the existing gloss-annotated training corpus and an LLM for sentence generation: per-gloss clips are extracted from training videos via CTC forced-alignment, novel gloss--sentence pairs are generated by a corpus-anchored LLM, and synthetic sequences are assembled through random sentence sampling and clip assignment. 
The resulting synthetic RGB video--text pairs are architecture-agnostic at the downstream training stage and can be consumed directly by RGB-based SLT models, or converted into pose or feature representations by pipelines that derive such inputs from video. Sincan et al.\ re-evaluated five recent gloss-free methods under strictly identical conditions; the largest verified gain over the GFSLT-VLP baseline was only 0.98 BLEU-4. Our augmentation, applied within the same framework, achieves \textbf{+2.92 BLEU-4} without any change to architecture or training protocol. 
We further identify that synthetic data harms vision-language
pretraining despite improving its objectives, and that optimising
clip transitions for visual smoothness is counter-productive under
L2-based criteria; we propose that abrupt boundaries may act as
a form of implicit regularisation.
Code is available at \url{https://github.com/robizso/slt-datagen}.
\end{abstract}

\keywords{data augmentation \and gloss-free sign language translation \and large language models \and sign stitching \and synthetic training data \and vision-language pretraining}


\section{Introduction}
\label{sec:introduction}

Sign languages are natural visual languages and a
cornerstone of Deaf culture, used not only by Deaf and
hard-of-hearing individuals but by entire signing communities.
They convey meaning through combinations of handshape, movement,
location, facial expression, and body posture, and vary considerably
in grammar and lexicon.
The central task addressed in this paper is
\textbf{Sign Language Translation} (SLT), which maps sign video to
spoken-language text. SLT has seen rapid progress through
advances in visual representation learning and large pretrained
language models.

The data landscape for SLT splits into two broad regimes.
In the \textit{gloss-annotated} regime, signing is transcribed
into gloss tokens and paired with a spoken-language translation
with tight clip-level alignment; such corpora are genuinely small,
as gloss annotation is expert-intensive~\cite{sincan2025gloss}.
In the \textit{gloss-free} regime, no sign-level annotation is
required, enabling larger-scale collection. This regime includes
curated datasets with tight alignment such as How2Sign~\cite{duarte2021how2sign}
and weakly aligned broadcast or web corpora that reach thousands
of hours at scale~\cite{li2025unisign} but yield modest pretraining
gains due to loose alignment, interpreter inconsistencies, and
domain mismatch. Despite these advances, high-quality parallel
video--text pairs for fine-tuning remain scarce, limiting
generalisation on long-tail vocabulary and unseen constructions,
and motivating data-centric approaches to improving SLT.

Augmenting sign language data is non-trivial.
Unlike text augmentation, sign language augmentation must contend
with the visual modality: realistic signing requires continuous
articulation with natural co-articulation between signs.
Feature-level methods~\cite{zhou2021signbt} stitch sign embeddings
but are tied to a specific encoder.
Skeleton-level methods~\cite{walsh2025slp,joshi2025posestitch}
produce pose sequences but are limited to pose-based architectures
unless combined with a rendering step that has yet to be explored.
We propose a simpler approach: directly stitching RGB video clips.
Throughout this paper, we use \emph{augmentation} to refer to the
overall corpus expansion process, and \emph{synthetic data} to refer
to the generated video--text training samples produced by the pipeline.

Our pipeline extracts per-gloss clips from existing training videos
via CTC forced-alignment segmentation~\cite{zuo2024online}, generates
novel gloss--sentence pairs with a corpus-anchored LLM, and assembles
synthetic sequences through random sentence sampling and clip assignment.
The pipeline requires no external video data or generative video models,
relying only on the gloss annotations available in the existing training
corpus and an LLM for sentence generation,
and produces standard RGB video--text pairs that are architecture-agnostic at the downstream training stage, consumable directly by RGB-based SLT models or convertible into pose and feature representations by pipelines that derive such inputs from video.
A contamination analysis confirms that generated gloss sequences are
no closer to the dev and test sets than real training sentences,
ruling out evaluation leakage as a confound.

The methodological context of our evaluation is important.
Sincan et al.~\cite{sincan2025gloss} reproduced five recent gloss-free
SLT methods under strictly identical conditions and found that most
reported improvements in the literature largely disappear under
controlled evaluation, with the largest verified gain over their
reproduced GFSLT-VLP baseline being only
21.97~$\to$~22.95 (\textbf{+0.98 BLEU-4}).
Our reproduction of the same baseline scores 21.38, slightly below
Sincan et al.'s 21.97, a discrepancy we attribute to mBART vocabulary
selection details not fully specified in the published codebase; our augmentation raises this to
21.38~$\to$~24.30 (\textbf{+2.92 BLEU-4}), without any
change to architecture, training objective, or hyperparameters.
To verify that the gain is not an artefact of increased gradient steps,
we evaluate a double-sampled real data baseline matched for training
computation,
which scores 21.17, statistically indistinguishable from the
real-only baseline of 21.38, confirming that additional gradient
steps alone do not explain the gain; our augmentation achieves
\textbf{+3.13 BLEU-4} over this compute-matched baseline
(21.17~$\to$~24.30).

Our main contributions are:
\begin{itemize}
    \item A scalable corpus augmentation pipeline combining CTC-based
    forced-alignment clip extraction, corpus-anchored LLM generation,
    and random sentence sampling and clip assignment, producing architecture-agnostic RGB video--text pairs consumable
    directly by RGB-based SLT models or convertible into pose and
    feature representations by pipelines that derive such inputs from video.

    \item Under the controlled conditions of Sincan et al.~\cite{sincan2025gloss},
    training with our augmented dataset yields
    \textbf{+2.92 BLEU-4} over our reproduced GFSLT-VLP baseline
    (21.38~$\to$~24.30), with no architectural change; a double-sampled real data baseline matched for training computation
    scores 21.17, statistically indistinguishable from the real-only
    baseline, confirming the gain is not an artefact of increased
    gradient steps (\textbf{+3.13 BLEU-4} over the doubled baseline).

    \item A counter-intuitive empirical finding that L2-based transition
    optimisation, despite producing visually smoother clip boundaries,
    is harmful for SLT fine-tuning; we propose that abrupt boundaries
    may act as implicit regularisation, encouraging the model to
    recognise signs from core visual content rather than temporal
    boundary context.
\end{itemize}


\section{Related Work}
\label{sec:related}

\subsection{Sign Language Translation}
\begin{figure}[t]
\centering
\includegraphics[width=0.6\columnwidth]{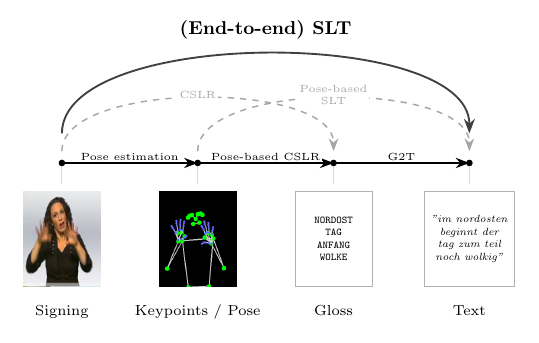}
\caption{The SLT pipeline. The bottom solid path shows the fully
  modularised pipeline: pose estimation extracts keypoints, which
  feed pose-based CSLR to produce a gloss sequence, followed by
  G2T translation to text.
  The top bold curve shows end-to-end SLT, mapping sign video
  directly to text.
  The dashed paths show common partial pipelines: CSLR alone
  (signing to gloss, skipping translation), and pose-based SLT
  (keypoints directly to text, skipping the gloss stage).}
\label{fig:slt_pipeline}
\end{figure}

Sign language translation (SLT) is the task of automatically converting sign
language video into spoken-language text.
As Fig.~\ref{fig:slt_pipeline} illustrates, two paradigms have emerged.
The \textit{gloss-based} paradigm uses continuous sign language recognition (CSLR) to produce a gloss sequence from video, which is then extended with a gloss-to-text (G2T) stage to produce the final translation.
The \textit{gloss-free} paradigm maps sign video directly to text, eliminating
gloss annotation. An important implementation dimension cuts across both paradigms:
rather than operating on RGB video, many CSLR and SLT systems use
body, hand, and face keypoint sequences as input, substantially
reducing computational cost while achieving competitive performance,
as MSKA~\cite{guan2025mska} demonstrates.

Camg\"{o}z et al.~\cite{camgoz2018neural} formulated SLT as a neural
sequence-to-sequence problem and released Phoenix-2014T, the benchmark
used throughout this work; the same group later introduced the Sign
Language Transformer~\cite{camgoz2020sign}, jointly training CSLR and
G2T with shared encoder weights and CTC supervision.
Chen et al.~\cite{chen2022twostream} proposed TwoStream-SLR, a dual-encoder
framework that processes RGB video and keypoint heatmaps in parallel through
bidirectional lateral connections and frame-level self-distillation, achieving
28.95 BLEU-4 on Phoenix-2014T.
Most recently, Guan et al.~\cite{guan2025mska} introduced MSKA, a multi-stream
keypoint attention network that decouples signing into body, hand, and face
streams modeled by pure self-attention without requiring manual graph-topology
design, reaching 29.03 BLEU-4, the current state of the art on this benchmark.
Recent work suggests that the G2T stage is no longer the limiting
factor in this pipeline, as large language models achieve near-human
gloss-to-English translation accuracy without any fine-tuning~\cite{alafandi2025llm},
pointing to video-to-gloss recognition as the primary remaining bottleneck.
Gloss annotation is expert-intensive, requiring approximately one hour per
90 seconds of video~\cite{sincan2025gloss}, which constrains dataset scale and
motivates the data augmentation approach of this paper.

The field has nonetheless shifted strongly toward gloss-free approaches due to
their lower annotation cost and greater scalability.
Zhou et al.~\cite{zhou2023gloss} introduced GFSLT-VLP, adapting CLIP-style
vision-language pretraining to sign language through contrastive and masked
language modeling objectives with a pretrained mBART decoder~\cite{tang2020multilingual}.
Its public implementation has become the de facto foundation for subsequent
work: SignCL~\cite{ye2024signcl} adds frame-level contrastive learning to reduce
representation density; C2RL~\cite{chen2025c2rl} combines cross-lingual
contrastive loss with lightweight translation pretraining; Sign2GPT~\cite{wong2024sign2gpt}
leverages LLM decoders with pseudo-gloss pretraining.
Under controlled, identical training conditions, however, performance gains from
these methodological contributions are modest~\cite{sincan2025gloss}.
The most significant recent advance is Li et al.~\cite{li2025unisign}, whose
Uni-Sign framework demonstrates that large-scale pretraining can close the
gloss-based gap. Pretrained on CSL-News, a 1,985-hour Chinese Sign Language broadcast corpus, Uni-Sign achieves 26.36 BLEU-4 on CSL-Daily, surpassing all gloss-based baselines on that benchmark.
This result confirms that pretraining data scale, rather than gloss supervision
alone, is a central driver of translation performance, and motivates data-centric
contributions such as ours.

The work of Sincan et al.~\cite{sincan2025gloss} is the most important reference
for our evaluation.
They re-implement five gloss-free methods: GFSLT-VLP~\cite{zhou2023gloss},
SignCL~\cite{ye2024signcl}, Sign2GPT~\cite{wong2024sign2gpt},
FLa-LLM~\cite{chen2024flallm}, and C2RL~\cite{chen2025c2rl}, in a unified
codebase under identical conditions, standardizing backbone, preprocessing,
training schedule, random seeds, and evaluation protocol.
Their key finding is that most reported improvements in the literature diminish
substantially under controlled evaluation, with the largest verified gain over
the GFSLT-VLP baseline being only \textbf{+0.98 BLEU-4} (21.97~$\to$~22.95).
Their open-source codebase and evaluation protocol serve as the direct foundation
for all our experiments, making our results directly comparable to theirs.

\subsection{Sign Segmentation}

Producing per-gloss video clips from continuous signing without
time-aligned annotations is a prerequisite for our augmentation pipeline.
We adopt the CTC forced-alignment segmentor of Zuo et al.~\cite{zuo2024online},
which, given the ground-truth gloss sequence, estimates per-sign frame
boundaries without requiring manual timing annotation.
Zuo et al.\ report 93.4\% segmentation accuracy in a manual evaluation on
Phoenix-2014T, providing a strong but imperfect signal; the remaining
boundary errors form a recoverable noise floor in our assembled synthetic
sequences.
Time-aligned gloss annotations, where available in future corpora, would
directly reduce this noise and represent the primary bottleneck for further
improvement of our approach.

\subsection{Data Augmentation for Sign Language}

Sign language training data can be augmented at many levels, from
simple visual perturbations (colour jitter, temporal resampling,
random cropping) to generating entirely new training sequences.
This section focuses on the latter: methods that synthesize new
gloss--sentence pairs, generate new signing sequences in feature,
skeleton, or video space, or apply steps of the sign language
production pipeline to create synthetic training data.
We organize these methods by the modality in which synthesis occurs:
\emph{text-side} generation of gloss--sentence pairs;
\emph{feature-level} sequence generation by stitching in embedding
space;
\emph{skeleton-level} sequence generation on pose sequences; and
\emph{photo-realistic rendering} to produce appearance-diverse video.
Table~\ref{tab:textgloss} summarizes text-side methods;
Table~\ref{tab:visual_aug} places all visual augmentation results in a
common frame.

\subsubsection{Text-Side Gloss--Sentence Synthesis}

All visual augmentation pipelines depend on a supply of gloss--sentence
pairs.
Three strategies appear in the literature.
\emph{Rule-based synthesis} requires no parallel data:
Moryossef et al.~\cite{moryossef2021data} derive pseudo-glosses from
monolingual text via lemmatization, function-word deletion, and
word-order permutation.
\emph{Model-based synthesis} trains a text-to-gloss (T2G) model on
available pairs and applies it to large monolingual corpora.
Zhou et al.~\cite{zhou2021signbt} use this in SignBT, generating
pairs more than $30\times$ the annotated corpus size; by training
T2G models of varying quality and measuring the resulting SLT
performance, they show that SLT results improve monotonically
with T2G accuracy, identifying it as the key bottleneck.
Abdullah et al.~\cite{abdullah2025gloss} replace the trained model
with GPT-4o prompted from approximately 100 annotated seed pairs.
\emph{In-corpus permutation} (Walsh et al.~\cite{walsh2025slp})
generates synthetic pairs by randomly permuting the gloss order
of existing training sequences, producing new gloss--sentence
combinations without introducing new vocabulary or content.
The gain over unpermuted stitched sequences is marginal (up to
$+0.21$ BLEU-4), suggesting that grammatical reordering alone
adds limited diversity.

Our approach belongs to the model-based category: each LLM call
receives the full Phoenix vocabulary and 500 randomly sampled
in-corpus examples as context, anchoring generation within the
Phoenix domain and weather-broadcast register.
In our ablation study, we additionally compare against a simpler
baseline that reuses ground-truth gloss sequences without permutation,
analogous to the unpermuted condition of Walsh et al. This provides
a controlled comparison that isolates the contribution of genuinely
new LLM-generated content from simply varying signer and clip
assignments on existing pairs.
Table~\ref{tab:textgloss} summarizes the text-side methods.

\begin{table}[t]
\centering
\caption{Text-side gloss--sentence synthesis methods. Methods differ in
  information source, degree of corpus anchoring, and annotation
  requirements.}
\label{tab:textgloss}
\small
\begin{tabular}{p{3.0cm}p{6.0cm}p{5.55cm}}
\toprule
\textbf{Method} & \textbf{Strategy} & \textbf{Resource} \\
\midrule
\multicolumn{3}{l}{\textit{Rule-based}} \\
\midrule
Moryossef et al.~\cite{moryossef2021data}
  & Lemmatize + delete function words + reorder
  & Tagesschau (DE); ASLG-PC12 (EN) \\
\midrule
\multicolumn{3}{l}{\textit{Model-based}} \\
\midrule
SignBT~\cite{zhou2021signbt}
  & Trained T2G applied to monolingual corpus
  & Wikipedia + weather (DE); WebText (ZH) \\
\midrule
Abdullah et al.~\cite{abdullah2025gloss}
  & LLM few-shot from $\sim$100 annotated seeds
  & Annotated BdSL seed pairs \\
\midrule
\textbf{This work}
  & LLM anchor-prompted generation
  & Phoenix vocab + 500 in-corpus examples \\
\midrule
\multicolumn{3}{l}{\textit{In-corpus permutation}} \\
\midrule
Walsh et al.~\cite{walsh2025slp}
  & Gloss-order permutation of existing pairs
  & Phoenix-2014T train split \\
\bottomrule
\end{tabular}
\end{table}

\subsubsection{Feature-Level Augmentation}

Feature-level augmentation generates new training sequences by
stitching sign feature clips in embedding space rather than at the
pixel or pose level.
The synthetic data is tied to a specific visual encoder, since the
same encoder must be used both to build the feature bank and at
inference time.

Zhou et al.~\cite{zhou2021signbt} introduce SignBT, building
a gloss-to-sign feature bank by aligning video features to gloss
boundaries via CTC forced alignment over the training corpus.
Synthetic training pairs are generated by back-translating large
monolingual corpora into pseudo-gloss sequences and splicing the
corresponding feature clips, with random selection providing
epoch-level diversity.
The resulting corpus yields $+2.64$ BLEU-4 on Phoenix-2014T and $+8.15$
on CSL-Daily, where the larger vocabulary amplifies the benefit.

\subsubsection{Skeleton-Level Augmentation}

Skeleton-level augmentation operates directly on pose coordinate
sequences, requiring a pose estimation front-end and a gloss-indexed
pose dictionary, either mined from the training corpus or drawn from
an external isolated-sign dataset.
A key advantage of the pose domain is that synthetic sequences can be
made kinematically more plausible than pixel-level stitching allows.
Because keypoint coordinates occupy a continuous, structured space,
stitch boundaries can be smoothed by modeling the velocity and
acceleration profiles of joints across transitions, signer-specific
proportions can be normalized to a canonical scale, and co-articulation
effects can be approximated analytically. None of these transformations have a direct equivalent at the pixel level.

Walsh et al.~\cite{walsh2025slp} apply skeleton stitching for SLT
data augmentation.
They build a gloss-pose dictionary from an external collection of
isolated DGS signs~\cite{hanke2020dgs}, covering the full Phoenix-2014T
gloss vocabulary, and pre-train GFSLT-VLP on the resulting synthetic
sequences, raising BLEU-4 from 11.32 to 13.68 on Phoenix-2014T.
Joshi et al.~\cite{joshi2025posestitch} demonstrate skeleton stitching
in a fully gloss-free, external-dictionary setting.
Drawing on WLASL~\cite{li2020wlasl} (ASL) and CISLR~\cite{joshi2022cislr}
(ISL) isolated sign datasets as pose banks, they generate sentences via
linguistic templates, stitch corresponding poses, and pre-train a
transformer for gloss-free SLT. This raises BLEU-4 from 1.97 to 4.56 on
How2Sign and from 0.55 to 3.43 on iSign, benchmarks that carry
neither gloss labels nor an in-corpus dictionary, making external sign
banks the only viable option.

\subsubsection{Photo-Realistic Rendering}

Walsh et al.~\cite{walsh2025slp} evaluate signer appearance augmentation
as a separate strategy from skeleton stitching, targeting video-based
rather than pose-based SLT architectures.
They render new signer appearances from the original Phoenix-2014T
skeleton poses using two models.
SignGAN~\cite{saunders2022signing} generates video from pose and a
style image via adversarial training, but is prone to artefacts caused
by noise in the conditioning.
SignSplat~\cite{ivashechkin2025signsplat} attaches 3D Gaussian
primitives to an SMPL-X human mesh with physiological joint
constraints, producing fewer artefacts and a larger downstream gain
($+2.26$ vs.\ $+1.24$ BLEU-4 over the baseline).
Walsh et al.\ identify as a direction for future work the combination of rendered appearance with synthetically stitched motion to generate entirely new video sequences from novel viewpoints.

\subsubsection{Positioning of This Work}

Table~\ref{tab:visual_aug} places our approach in the context of prior
augmentation results, introducing RGB video stitching as a distinct
category alongside feature-level, skeleton-level, and rendering-based
methods.
Skeleton-level methods are limited to pose-based SLT models; extending
them to RGB-based architectures would require an additional rendering
step, a combination that has yet to be explored.
Our pipeline operates directly on raw video without pose estimation,
encoder coupling, or appearance modelling, making it applicable to any
RGB-based SLT framework.
Absolute BLEU-4 scores in Table~\ref{tab:visual_aug} should be treated
with caution: results differ in input modality, baseline model,
dataset split, and scoring convention across papers, making direct
numerical comparison unreliable.

\begin{table*}[t]
\centering
\caption{Visual augmentation results on sign language translation
  benchmarks. \textbf{P}~=~Phoenix-2014T, \textbf{C}~=~CSL-Daily,
  \textbf{H}~=~How2Sign, \textbf{iS}~=~iSign.
  \textbf{V2T}~=~video-to-text; \textbf{P2T}~=~pose-to-text.
  Baseline and augmented BLEU-4 are not directly comparable across
  modalities. }
\label{tab:visual_aug}
\begin{tabular}{llllcc}
\toprule
\textbf{Method} & \textbf{Modality} & \textbf{Sign source} &
\textbf{Bench.} & \textbf{Base B4} & \textbf{Aug B4 ($\boldsymbol{\Delta}$)} \\
\midrule
\multicolumn{6}{l}{\textit{Feature-level (V2T)}} \\
\midrule
SignBT~\cite{zhou2021signbt}
  & Feature stitching & Same corpus
  & P & 21.68 & 24.32 {\scriptsize($+$2.64)} \\
  & & & C & 13.19 & 21.34 {\scriptsize($+$8.15)} \\
\midrule
\multicolumn{6}{l}{\textit{Skeleton-level (P2T)}} \\
\midrule
Walsh et al.~\cite{walsh2025slp}
  & Skeleton stitching & Same corpus + DGS~\cite{hanke2020dgs}
  & P & 11.32 & 13.68 {\scriptsize($+$2.36)} \\
Joshi et al.~\cite{joshi2025posestitch}
  & Skeleton stitching & WLASL~\cite{li2020wlasl}
  & H  & 1.97 & 4.56 {\scriptsize($+$2.59)} \\
  & & CISLR~\cite{joshi2022cislr}
  & iS & 0.55 & 3.43 {\scriptsize($+$2.88)} \\
\midrule
\multicolumn{6}{l}{\textit{Photo-realistic rendering from ground-truth pose (V2T)}} \\
\midrule
Walsh + SignGAN~\cite{walsh2025slp,saunders2022signing}
  & GAN rendering & Real pose
  & P & 19.53 & 20.77 {\scriptsize($+$1.24)} \\
Walsh + SignSplat~\cite{walsh2025slp,ivashechkin2025signsplat}
  & Gaussian splatting & Real pose
  & P & 19.53 & 21.79 {\scriptsize($+$2.26)} \\
\midrule
\multicolumn{6}{l}{\textit{RGB video stitching, this work (V2T)}} \\
\midrule
\textbf{Ours}
  & \textbf{RGB video stitching} & Same corpus
  & P & 21.38 & \textbf{24.30} {\scriptsize($\mathbf{+2.92}$)} \\
\bottomrule
\end{tabular}
\end{table*}


\section{Method}
\label{sec:method}

\subsection{Overview}
Our pipeline consists of three stages: (1) per-gloss video clip
extraction via CTC forced-alignment segmentation; (2) LLM-guided
corpus-anchored generation of gloss--sentence pairs; and (3)
random sentence sampling, clip assignment, and video assembly.

\subsection{Per-Gloss Video Clip Extraction}
\label{sec:method:segmentation}

Phoenix-2014T provides gloss sequences as ground-truth labels but
does not include time-aligned boundary annotations.
We adopt the CTC forced-alignment segmentor of Zuo et
al.~\cite{zuo2024online}, which conditions a pre-trained sign
recognizer on the ground-truth gloss sequence to estimate frame-level
boundaries without manual annotation.
Given a video $V = (v_1, \ldots, v_T)$ and gloss sequence
$G = (g_1, \ldots, g_N)$, the aligner produces start and end frame
estimates $(s_i, e_i)$ for each gloss $g_i$.
Zuo et al.\ report 93.4\% segmentation accuracy in a manual
evaluation on Phoenix-2014T; the remaining boundary errors form a
recoverable noise floor in the assembled synthetic sequences.
We apply this segmentor to all 7,096 training videos, storing clips
as index references into the original LMDB database with no
additional disk space.

\subsection{LLM-Guided Gloss--Sentence Pair Generation}
\label{sec:method:generation}

Naive prompting without domain grounding produces vocabulary drift
and German text inconsistent with the weather-broadcast register.
We address this through corpus anchoring: each API call provides the
model with the complete Phoenix gloss vocabulary, a random sample of
corpus examples as context, and one or more \emph{anchor pairs}
drawn from the training set.
The model is instructed to generate new pairs that resemble the
anchor group in structure and register, but vary the content.

To ensure coverage of the full training distribution, anchors are
drawn using a least-used selection strategy: we track how many times
each training sentence has served as an anchor and always prefer
the least-frequently used.
This encourages diversity in the generated pairs while preventing
the model from drifting outside the Phoenix domain.

Generated sequences are validated against the Phoenix vocabulary
and filtered to remove exact matches with training corpus sentences,
strict subsequences of corpus sentences, and duplicates within the
generated pool, ensuring that all retained pairs constitute genuinely
new linguistic content.

\subsection{Sentence Sampling, Clip Assignment, and Video Assembly}
\label{sec:method:selection}

From the generated pool, $N$ sentences are sampled uniformly at random,
and each is assembled into a synthetic video by concatenating per-gloss
clips from the training database.

\paragraph{Signer and clip selection.}
Each sentence is assigned to a single signer drawn uniformly at random
from those who have recorded clips for every gloss in the sequence.
For LLM-generated sentences, no original signer exists; the signer is
therefore sampled uniformly from all eligible signers.
In the ablation condition that reuses original Phoenix training
sentences (Section~\ref{sec:ablation}), the original signer is
excluded to avoid reconstructing the original training video.
Clip instances for each gloss are then selected uniformly at random
from the available occurrences for the assigned signer.
As an optional post-processing step, clips can be re-selected to
minimise the L2 distance between consecutive boundary frames,
producing visually smoother transitions; however, we find this
consistently hurts translation metrics and therefore exclude it from
our primary pipeline.

\begin{figure*}[t]
\centering
\includegraphics[width=\linewidth]{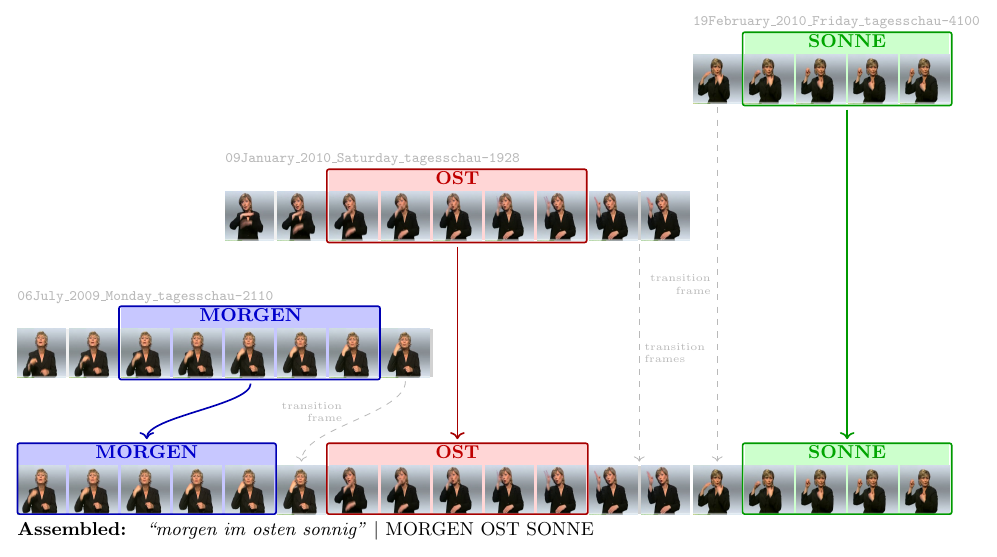}
\caption{Video assembly illustrated for the synthetic sample
  \textit{``morgen im osten sonnig''} / MORGEN OST SONNE\@.
  Each source row shows sampled frames from a Phoenix-2014T training
  video; coloured rectangles mark the core gloss clip extracted by
  CTC forced-alignment segmentation, and grey regions are transition
  frames drawn from the inter-gloss gap.
  Thick arrows show the origin of each core clip in the assembled
  sequence; thin dashed arrows show the origin of transition frames,
  which may come from the tail of the preceding clip (after OST) or
  the head of the following clip (before SONNE).
  The assembled row concatenates all clips in gloss order
  (6\,+\,8\,+\,8 = 22 frames total; not all frames shown).}
\label{fig:assembly}
\end{figure*}

\paragraph{Video assembly.}
The synthetic video is assembled by concatenating the selected frame
subsequences, read directly from the existing Phoenix LMDB database
via the index references stored at the clip extraction stage;
Fig.~\ref{fig:assembly} illustrates this process on a concrete example.
At each boundary between consecutive glosses, a transition frame
budget is sampled from the empirical distribution of inter-gloss
gaps observed in the real Phoenix training set, capped at the 95th
percentile to avoid unrealistically long pauses.
This budget is split between the tail frames of the preceding clip
and the head frames of the following clip, proportionally and subject
to the frames available in each source video.
No additional disk space is required beyond the index files; at
training time, frames are loaded on-the-fly and pass through the
same preprocessing pipeline as real samples.

\section{Experimental Setup}
\label{sec:experiments}

\subsection{Dataset}
We evaluate on Phoenix-2014T~\cite{camgoz2018neural}, containing
7,096 training, 519 development, and 642 test samples of DGS weather
broadcast signing with German translations.

\subsection{Augmentation Conditions}
All primary experiments use 7,000 synthetic samples combined with the
7,096 real training samples (1$\times$ real). Primary results are
reported at 200 epochs over three seeds. Exploratory analyses
(synthetic VLP pretraining and transition optimisation) are run with
a single seed at 75 epochs for computational efficiency; all such
runs are clearly marked in the results.

\subsection{Implementation Details}

\subsubsection{Generation}
Gloss--sentence pairs are generated using \texttt{gpt-5.4-mini} (OpenAI,
April 2026) with API default sampling parameters (temperature 1.0,
top-p 1.0), producing 10 pairs per API call with 500 randomly sampled
in-corpus examples as context. The selected 7k synthetic gloss--sentence
list is released alongside the code at https://github.com/robizso/slt-datagen
to support reproducibility and auditing.
Table~\ref{tab:gen_stats} summarises the full generation pipeline statistics.

\begin{table}[t]
\centering
\caption{Synthetic corpus generation statistics.}
\label{tab:gen_stats}
\small
\begin{tabular}{lr}
\toprule
\textbf{Quantity} & \textbf{Value} \\
\midrule
Candidate pairs generated          & 124,219 \\
Removed: exact gloss match         & 1,428 \\
Removed: gloss subsequence         & 3,689 \\
After gloss-level filtering        & 119,102 \\
After deduplication                & 115,547 \\
Removed: no feasible signer        & 4,485 \\
With feasible signer assignment    & 111,062 \\
Selected training subset           & 7,000 \\
\midrule
Mean gloss length — real           & 7.8 tokens \\
Mean gloss length — synthetic      & 7.9 tokens \\
Gloss vocab coverage — real        & 1,085 / 1,085 \\
Gloss vocab coverage — synthetic   & 627 / 1,085 \\
\bottomrule
\end{tabular}
\end{table}

\subsubsection{Training}
All experiments follow the GFSLT-VLP framework~\cite{zhou2023gloss}
as reproduced by Sincan et al.~\cite{sincan2025gloss}, using a patched version of
their open-source codebase that supports synthetic label files,
available at https://github.com/robizso/sltbaselines-synthetic.
We use a vocabulary-trimmed mBART model with 2,215 tokens.
Training hyperparameters: batch size 8, SGD with momentum 0.9,
learning rate $10^{-2}$ with cosine annealing, weight decay $10^{-3}$.
Primary results are averaged over seeds 0, 42, and 100, consistent
with Sincan et al.~\cite{sincan2025gloss}; exploratory experiments use seed 0.
VLP pretraining: learning rate $5\times10^{-3}$, 80 epochs.
Hardware: single NVIDIA L40S (48 GB) or PRO 6000 Blackwell (96 GB).

\subsection{Evaluation Methodology}
\label{sec:eval}
We report BLEU-1 to BLEU-4~\cite{papineni2002bleu} and
ROUGE-L~\cite{lin2004rouge}.
Following Phoenix-2014T convention, a period token is appended to
both predictions and references.
BLEU is computed with sacrebleu~\cite{post2018call} using 13a
tokenization and exponential smoothing, consistent with Sincan et al.~\cite{sincan2025gloss}
(signature: \texttt{nrefs:1|case:mixed|eff:no|tok:13a|smooth:exp|version:2.6.0}).
ROUGE-L uses pycocoevalcap.
The \texttt{de\_DE} mBART language token is stripped from predictions
and references prior to scoring.

\subsection{Baseline Reproduction}
\label{sec:baseline}
Our reproduced baseline at 200 epochs is shown in
Table~\ref{tab:scoring} alongside the original GFSLT-VLP results and
Sincan et al.'s reproduction.
Our reproduction closely matches the original GFSLT-VLP paper, with
higher BLEU-1 (46.04 vs.\ 43.71) but similar BLEU-4 (21.38 vs.\
21.44), consistent with differences in mBART vocabulary selection
which is not fully specified in the published codebase.
Sincan et al.'s~\cite{sincan2025gloss} reproduction scores higher across all metrics;
their ROUGE-L in particular (46.87) differs substantially from both
the original paper (42.49) and our reproduction (41.69), suggesting
different scoring settings beyond the stated nlg-eval library.
To verify that our augmentation gain is not an artefact of increased
gradient steps, we additionally evaluate a double-sampled real data
baseline that matches the per-epoch computation of our augmented
setting. We treat our reproduced baseline as the reference point for
all comparisons in this paper.

\begin{table}[t]
\centering
\caption{Baseline reproduction on the Phoenix-2014T test set at
  200 epochs. $\pm$ values are std over 3 seeds.}
\label{tab:scoring}
\small
\begin{tabular}{lrrrrr}
\toprule
Method & B1 & B2 & B3 & B4 & RL \\
\midrule
Zhou et al.~\cite{zhou2023gloss}        & 43.71 & 33.18 & 26.11 & 21.44 & 42.49 \\
Sincan et al.~\cite{sincan2025gloss}      & 46.44 & 33.96 & 26.69 & 21.97 & 46.87 \\
\midrule
Reprod.\ (200 ep) & 46.04$\pm$0.27 & 33.54$\pm$0.30 & 26.08$\pm$0.27 & 21.38$\pm$0.24 & 41.69$\pm$0.27 \\
Real doubled  & 46.17$\pm$0.51 & 33.53$\pm$0.64 & 26.00$\pm$0.60 & 21.17$\pm$0.48 & 41.66$\pm$0.47 \\
\bottomrule
\end{tabular}
\end{table}


\section{Results}
\label{sec:results}

\subsection{Primary Results (200 Epochs)}
\label{sec:results:main}

Table~\ref{tab:main_results} summarizes the principal augmentation
results at 200 epochs, averaged over three seeds.
Augmenting SLT fine-tuning with synthetic data consistently improves
over the real-data baseline.
Our primary result, 7k synthetic samples at 200 epochs, achieves
mean BLEU-4 = 24.30 and ROUGE-L = 46.09, corresponding to a mean
improvement of $+2.92$ BLEU-4 over our reproduced baseline
(21.38~$\to$~24.30).
The double-sampled real data baseline, matched for training
computation, scores 21.17 BLEU-4, marginally \emph{below} the
real-only baseline of 21.38 and well within one standard deviation,
confirming that additional gradient steps provide no benefit and that
the \textbf{+2.92 BLEU-4} gain stems entirely from the synthetic
content.

\begin{table*}[t]
\centering
\caption{Main augmentation results on Phoenix-2014T test set at
200 epochs. $\pm$ values are std over 3 seeds.}
\label{tab:main_results}
\small
\begin{tabular}{lrrrrr}
\toprule
Setting & B1 & B2 & B3 & B4 & RL \\
\midrule
Baseline (real only)
& 46.04$\pm$0.27 & 33.54$\pm$0.30 & 26.08$\pm$0.27 & 21.38$\pm$0.24 & 41.69$\pm$0.27 \\
Baseline (real doubled)
& 46.17$\pm$0.51 & 33.53$\pm$0.64 & 26.00$\pm$0.60 & 21.17$\pm$0.48 & 41.66$\pm$0.47 \\
Ours
& \textbf{50.04}$\pm$0.64 & \textbf{37.30}$\pm$0.57 & \textbf{29.41}$\pm$0.43 & \textbf{24.30}$\pm$0.31 & \textbf{46.09}$\pm$0.36 \\
\midrule
$\Delta$ vs.\ real only
& +4.00 & +3.76 & +3.33 & +2.92 & +4.40 \\
$\Delta$ vs.\ real doubled
& +3.87 & +3.77 & +3.41 & +3.13 & +4.43 \\
\bottomrule
\end{tabular}
\end{table*}

\subsection{Exploratory Analyses (75 Epochs)}
\label{sec:results:explore}

The following experiments are conducted at 75 epochs with a single
seed to enable broader exploration of augmentation behaviour under
practical computational constraints.
These analyses investigate the effects of synthetic VLP pretraining
and L2 transition optimisation.

\subsubsection{Synthetic VLP Pretraining Data}
\label{sec:results:vlp}

Table~\ref{tab:vlp} compares the effect of introducing synthetic data
during the VLP pretraining stage versus only at the SLT fine-tuning
stage. The real-VLP rows (R/R and R/R+S) are included for reference;
the computationally fair comparison for the fine-tuning condition is
discussed in Section~\ref{sec:results:main}.
The key finding is in the synthetic-VLP rows: introducing synthetic
data during VLP pretraining causes performance to collapse to
approximately 11 BLEU-4, regardless of whether synthetic data is
also used during SLT fine-tuning.

Interestingly, the synthetic-VLP checkpoints achieve lower
pretraining loss on the real Phoenix development set and better SLT
initialisation cross-entropy, yet exhibit substantially larger
gradient norms during SLT fine-tuning and converge much more slowly.
We attribute this to the unnatural frame transitions introduced by
stitched sequences: the visual encoder learns representations that
fit synthetic pretraining objectives but fail to generalise to real
continuous signing.

\begin{table}[t]
\centering
\caption{Effect of synthetic data at the VLP pretraining and SLT
fine-tuning stages on Phoenix-2014T test set (75 epochs, single seed).
R = real data only; R+S = real + synthetic data.}
\label{tab:vlp}
\small
\begin{tabular}{llrrrrr}
\toprule
VLP & SLT & B1 & B2 & B3 & B4 & RL \\
\midrule
R   & R   & 43.68 & 30.79 & 23.44 & 18.95 & 38.44 \\
R   & R+S & 49.57 & 36.52 & 28.63 & 23.56 & 45.46 \\
\midrule
R+S & R   & 33.97 & 20.74 & 14.74 & 11.38 & 28.43 \\
R+S & R+S & 33.47 & 20.67 & 14.51 & 11.19 & 27.65 \\
\bottomrule
\end{tabular}
\end{table}

\subsubsection{L2 Transition Optimisation}
\label{sec:results:l2}

A natural hypothesis is that smoother visual transitions between
stitched clips would produce more realistic synthetic sequences and
therefore improve translation quality.
We test this using two conditions at the 7k augmentation scale.
The first re-selects clip instances to minimise the L2 distance
between consecutive boundary frames while keeping the same sentence
pool fixed.
The second additionally filters candidate sentences based on the
smoothness of their achievable transitions.
All three conditions use the same number of synthetic samples and
thus the same gradient steps per epoch, making this a
computationally fair comparison.

\begin{table}[t]
\centering
\caption{Effect of L2 transition optimisation on Phoenix-2014T
test set (7k synthetic samples, 75 epochs, single seed).
I.~=~instance-level optimisation;
S.~=~sentence-level optimisation.}
\label{tab:l2}
\small
\begin{tabular}{lrrrrr}
\toprule
Setting & B1 & B2 & B3 & B4 & RL \\
\midrule
Random        & 49.57 & 36.52 & 28.63 & 23.56 & 45.46 \\
L2 (I.)       & 48.58 & 35.66 & 27.76 & 22.63 & 45.00 \\
L2 (I.+S.)    & 48.18 & 35.18 & 27.38 & 22.50 & 44.14 \\
\bottomrule
\end{tabular}
\end{table}

Table~\ref{tab:l2} shows that both L2-optimised conditions
underperform random selection, with losses of $-0.93$ and
$-1.06$ BLEU-4 respectively.
Sentence-level optimisation biases selection toward gloss sequences
whose clips happen to admit smooth transitions, reducing linguistic
diversity for a criterion unrelated to semantic content.
Instance-level optimisation preserves sentence diversity but reduces
clip-level diversity by repeatedly selecting the same smoothest
instances for each gloss.

These results suggest that, at least under L2-based transition
optimisation, visual smoothness at clip boundaries does not benefit
SLT fine-tuning and may in fact hurt it.
We propose that abrupt transitions may act as a form of implicit
regularisation, encouraging the model to recognise signs from core
visual content rather than temporal boundary cues; a targeted
experiment to verify this mechanism is beyond the scope of this paper.

\section{Ablation Study}
\label{sec:ablation}
To verify the necessity of LLM-generated content, we evaluate a
condition in which the generated gloss--sentence pairs are replaced
by the original Phoenix training sentences, while keeping random
clip assignment unchanged. Results are averaged over three seeds at 200 epochs.

\begin{table*}[t]
\centering
\caption{Effect of LLM-generated content on Phoenix-2014T test set
(7k samples, 200 epochs). $\pm$ values denote standard deviation over three seeds.}
\label{tab:ablation}
\small
\begin{tabular}{lrrrrr}
\toprule
Setting & B1 & B2 & B3 & B4 & RL \\
\midrule
Baseline (real only)
    & 46.04$\pm$0.27
    & 33.54$\pm$0.30
    & 26.08$\pm$0.27
    & 21.38$\pm$0.24
    & 41.69$\pm$0.27 \\
w/o LLM generation
    & 44.09$\pm$0.14
    & 31.63$\pm$0.12
    & 24.47$\pm$0.24
    & 20.04$\pm$0.26
    & 39.76$\pm$0.18 \\
\textbf{Ours}
    & \textbf{50.04}$\pm$0.64
    & \textbf{37.30}$\pm$0.57
    & \textbf{29.41}$\pm$0.43
    & \textbf{24.30}$\pm$0.31
    & \textbf{46.09}$\pm$0.36 \\
\bottomrule
\end{tabular}
\end{table*}

The gap of $-4.26$ BLEU-4 relative to our full pipeline strongly
suggests that LLM-generated content is not merely additive but
load-bearing.
Without it, performance falls \emph{below} the no-augmentation
baseline ($20.04$ vs.\ $21.38$ BLEU-4), with ROUGE-L also
dropping below baseline ($39.76$ vs.\ $41.69$).
Reassigning original training sentences to different signers and clip
instances introduces visual discontinuities at gloss boundaries
without compensating linguistic diversity: the model is exposed to
unfamiliar appearance variation for sequences whose translations it
has already seen, yielding no new learning signal while adding noise.
The result establishes that the visual variation introduced by
alternative-signer stitching is insufficient on its own: novel
gloss--sentence content is the primary driver of the augmentation
gain.


\section{Analysis}
\label{sec:analysis}

\subsection{Why Does Synthetic SLT Data Help?}
The ablation results point to a single dominant mechanism:
\textbf{linguistic diversity}.
LLM-generated sentences are not merely additive but essential.
Reassigning original training sentences to different signers and
clip instances, without introducing novel gloss--sentence pairs,
reduces performance below the no-augmentation baseline
(20.04 vs.\ 21.38 BLEU-4), falling 1.34 BLEU-4 below the real-only
baseline and 4.26 BLEU-4 below the full synthetic pipeline.
Visual variation alone is insufficient; novel linguistic content
is the primary driver of the augmentation gain.

\subsection{Why Does Synthetic VLP Data Hurt?}
The VLP stage optimises contrastive alignment between visual and
textual representations.
Stitched sequences introduce artificial visual discontinuities at
segment boundaries. The visual encoder can fit these during VLP, as
shown by lower pretraining loss on the real development set, but the
learned features do not generalise well to real continuous signing.
The doubled gradient norm at SLT initialisation with synthetic VLP
checkpoints indicates a fundamentally different optimisation landscape,
consistent with the finding of Ye et al.~\cite{ye2023xmda} that sign
video features and gloss embeddings occupy substantially different
distributions in end-to-end SLT models.

\subsection{Contamination and Novelty Analysis}
\label{sec:contamination}

To verify that the LLM does not reproduce evaluation content, we compare
the gloss sequences and German sentences of our 7k synthetic set against
the Phoenix-2014T train, dev, and test splits. Dev and test sets were
not used at any stage of generation or selection.

Table~\ref{tab:contamination} reports max-BLEU-4 of each gloss sequence
against its closest match in dev and test, computed on the full synthetic
set, alongside the same metric for real training sentences.
Zero exact gloss duplicates were found against any split, confirming
that the pipeline's gloss-level filtering is effective.
Synthetic gloss sequences show a consistently lower mean max-BLEU-4
against both dev (15.70 vs.\ 20.20) and test (16.18 vs.\ 21.25) than
real training sentences, confirming that the synthetic set is no closer
to the evaluation sets than the training data itself.

\begin{table}[t]
\centering
\caption{Contamination analysis: max-BLEU-4 of gloss sequences against
  Phoenix-2014T dev and test splits. Lower synthetic scores confirm
  no evaluation leakage. Exact duplicate check and max-BLEU-4 computed
  on the full 7k synthetic set.}
\label{tab:contamination}
\small
\begin{tabular}{lcccc}
\toprule
& \multicolumn{2}{c}{\textbf{vs.\ Dev}} & \multicolumn{2}{c}{\textbf{vs.\ Test}} \\
\cmidrule(lr){2-3} \cmidrule(lr){4-5}
& Mean & Max & Mean & Max \\
\midrule
Exact gloss duplicates & \multicolumn{2}{c}{0} & \multicolumn{2}{c}{0} \\
\midrule
Real train             & 20.20 & 100.00 & 21.25 & 100.00 \\
Synthetic              & 15.70 &  80.91 & 16.18 &  84.65 \\
\bottomrule
\end{tabular}
\end{table}

We additionally audit German target sentences, since the translation
is the supervised objective in SLT\@.
German sentence exact matches occur at rates of 0.59\% (3/509) against
dev and 1.27\% (8/630) against test, computed over the unique German
sentences in each split (509 and 630 respectively). The real
Phoenix-2014T training set itself overlaps at 2.95\% (15/509) and
3.17\% (20/630), more than twice the synthetic rate.
The matches arise exclusively from universal
broadcast phrases such as fixed greetings and date announcement
templates that appear across all splits regardless of source.
In all cases the gloss sequences differ, confirming that the supervised
video--text pairs are genuinely distinct.
Table~\ref{tab:sent_matches} shows representative examples.

\begin{table}[t]
\centering
\caption{Representative German sentence matches between synthetic and
  real splits. Gloss sequences differ in all cases, confirming
  independent generation. All matches are fixed broadcast phrases
  or date announcement templates.}
\label{tab:sent_matches}
\small
\begin{tabular}{p{3.2cm}p{3.2cm}p{3.2cm}}
\toprule
\textbf{German sentence} & \textbf{Synthetic gloss} & \textbf{Real gloss} \\
\midrule
guten abend liebe zuschauer
  & GUT ABEND ZUSCHAUER
  & ABEND LIEB ZUSCHAUER BEGRUESSEN \\
\midrule
und nun die wettervorhersage für morgen freitag den dritten dezember
  & JETZT WETTER MORGEN FREITAG DRITTE DEZEMBER WIE-AUSSEHEN
  & JETZT WETTER VORAUS SAGEN MORGEN FREITAG DRITTE DEZEMBER \\
\midrule
dabei bleibt es meist trocken
  & DAZU BLEIBEN MEISTENS TROCKEN
  & MEISTENS TROCKEN REGION REGEN negalp-KEIN \\
\bottomrule
\end{tabular}
\end{table}

\section{Conclusion}
\label{sec:conclusion}

We presented a corpus augmentation approach for sign language
translation that constructs synthetic training samples by segmenting
existing training videos into per-gloss clips via CTC forced-alignment,
generating novel gloss--sentence pairs with a corpus-anchored LLM, and
assembling synthetic sequences through random clip assignment.
The pipeline requires no external video data or generative video models,
relying only on the gloss annotations available in the existing training
corpus and an LLM for sentence generation, 
and produces standard RGB video--text pairs that are architecture-agnostic at the downstream training stage, directly consumable by RGB-based SLT models or convertible into pose and feature representations by pipelines that derive such inputs from video.

Sincan et al.~\cite{sincan2025gloss} established a rigorous controlled
evaluation framework in which the largest verified gain across five recent
gloss-free methods is only 0.98 BLEU-4 (21.97~$\to$~22.95).
Within this same framework, our augmentation achieves
$+2.92$ BLEU-4 over our reproduced 200-epoch baseline (21.38~$\to$~24.30),
with no change to the underlying architecture, training objective, or
hyperparameters. A double-sampled real data baseline, which matches our augmented
setting in gradient steps per epoch, scores 21.17, marginally
below the real-only baseline of 21.38 and well within one standard
deviation, confirming that the gain stems entirely from the
synthetic content, not from additional gradient steps.
A contamination analysis further confirms that
generated gloss sequences are no closer to the dev and test sets than
real training sentences, ruling out evaluation leakage as a confound.

Two additional findings emerge from our exploratory experiments.
First, in single-seed experiments at 75 epochs, introducing synthetic
stitched sequences during VLP pretraining substantially degraded
downstream SLT performance despite improving the pretraining objective,
suggesting that abrupt frame transitions prevent the visual encoder
from learning representations that generalise to real continuous signing.
Second, L2-based transition smoothing reduced translation quality in
our exploratory setting, suggesting that pixel-level boundary smoothness
is not a reliable proxy for useful augmentation quality; we propose
that abrupt boundaries may act as implicit regularisation, though a
targeted verification is beyond the scope of this paper.

Limitations and future directions include exploring larger augmentation
scales, improving segmentation quality through time-aligned gloss
annotations, and extending the approach to other sign language
benchmarks and languages.

\section*{Data and Code Availability}
The code for the augmentation pipeline, the selected 7,000
synthetic gloss--sentence pairs, and all evaluation scripts
are publicly available at \url{https://github.com/robizso/slt-datagen}.
A patched version of the GFSLT-VLP training codebase supporting
synthetic label files is available at
\url{https://github.com/robizso/sltbaselines-synthetic}.
Experiments use Phoenix-2014T~\cite{camgoz2018neural}, which is
publicly available at \url{https://www-i6.informatik.rwth-aachen.de/~koller/RWTH-PHOENIX/}.
Trained model checkpoints are not released at this time.


\section*{Acknowledgements}

The support of the GYORSÍTÓSÁV programme of the Hungarian National
Research, Development and Innovation Office is gratefully acknowledged.
Grant no.: 2023-1.1.2-GYORSÍTÓSÁV-2024-00019.


\appendix

\section{LLM Prompts}
\label{app:prompts}

\subsection{Generation System Prompt}
\label{app:gen_system}

When a length target range $[l_{\min}, l_{\max}]$ is active, the final line is replaced with \textit{``Target gloss sequence length: $l_{\min}$--$l_{\max}$ tokens (strict)''}.

\begin{small}
\begin{verbatim}
You are an expert in German Sign Language
(Deutsche Gebärdensprache, DGS).
Your task is to generate sentence--gloss
pairs in the style of German TV weather
broadcast signing.

Pairs consist of:
  - German spoken sentences (TV weather
    forecasts)
  - DGS gloss sequences: space-separated
    UPPERCASE tokens, topic-comment order:
    TIME → LOCATION → WEATHER EVENT
    → DEGREE/MODIFIER

Rules:
  • Use only tokens that occur exactly in the Phoenix vocabulary
  • Most gloss tokens are uppercase (REGEN, WIND, TEMPERATUR, KALT)
  • Negated tokens follow the corpus convention with a lowercase
    prefix (neg-WARM, neg-REGEN)
  • Omit articles, copulas, most prepositions
  • Typical length: 4–12 tokens
\end{verbatim}
\end{small}

\subsection{Generation User Prompt}
\label{app:gen_user}

\texttt{[VOCAB]} contains all 1,085 allowed gloss tokens separated by \texttt{|}. \texttt{[EXAMPLES]} contains 500 randomly sampled corpus pairs. \texttt{[ANCHORS]} contains 3 anchor pairs selected by least-used frequency.

\begin{small}
\begin{verbatim}
ALLOWED GLOSS VOCABULARY (1085 tokens):
[VOCAB]

EXAMPLE PAIRS (500 samples):
  SENTENCE: [german sentence]
  GLOSS:    [gloss sequence]
  ...

Focus especially on these seed pairs and
generate 10 new pairs that closely resemble
them:
  SENTENCE: [anchor sentence 1]
  GLOSS:    [anchor gloss 1]
  SENTENCE: [anchor sentence 2]
  GLOSS:    [anchor gloss 2]
  SENTENCE: [anchor sentence 3]
  GLOSS:    [anchor gloss 3]

Generate 10 NEW sentence–gloss pairs.
  - Use ONLY tokens from the allowed
    gloss vocabulary
  - Do NOT copy any example verbatim
  - Vary locations, temperatures, weather
    events, time references
\end{verbatim}
\end{small}

In single-anchor mode (group size 1), the anchor instruction is:
\begin{small}
\begin{verbatim}
Use the following as your template.
Generate 10 new pairs that closely mirror
its structure, gloss length, and complexity
— but with different content.

  SENTENCE: [anchor sentence]
  GLOSS:    [anchor gloss]

IMPORTANT: Match the gloss token count
(±1 token). Preserve the structural
pattern (TIME → LOCATION → WEATHER
→ DEGREE).
\end{verbatim}
\end{small}


\bibliographystyle{unsrt}
\bibliography{references}

\end{document}